\documentclass[11pt,a4paper]{article}
\usepackage[hyperref]{eacl2021}
\usepackage{times}
\usepackage{latexsym}
\usepackage{amsfonts}
\usepackage{xcolor}

\usepackage{graphicx}
\usepackage{subfigure}
\usepackage{tipa}
\usepackage{listings}
\definecolor{codegreen}{rgb}{0,0.6,0}
\definecolor{codegray}{rgb}{0.5,0.5,0.5}
\definecolor{codepurple}{rgb}{0.58,0,0.82}
\definecolor{backcolour}{rgb}{0.95,0.95,0.92}

\lstdefinestyle{mystyle}{
    backgroundcolor=\color{backcolour},   
    commentstyle=\color{codegreen},
    keywordstyle=\color{magenta},
    numberstyle=\tiny\color{codegray},
    stringstyle=\color{codepurple},
    basicstyle=\ttfamily\footnotesize,
    breakatwhitespace=false,         
    breaklines=true,                 
    captionpos=b,                    
    keepspaces=true,                 
    numbers=left,                    
    numbersep=5pt,                  
    showspaces=false,                
    showstringspaces=false,
    showtabs=false,                  
    tabsize=2
}

\usepackage{hyperref}
\hypersetup{
    colorlinks=true,
    linkcolor=blue,
    filecolor=magenta,      
    urlcolor=cyan,
}

\usepackage{microtype}

\aclfinalcopy 

\title{Contextual Text Embeddings for Twi}
\author{Paul Azunre$^{1*}$, Salomey Osei$^{2*}$, Salomey Afua Addo$^{3*}$,  Lawrence Asamoah Adu-Gyamfi$^{*}$, \\
\textbf{Stephen Moore$^{4*}$, Bernard Adabankah$^{5*}$, Bernard Opoku$^{6*}$, Clara Asare-Nyarko$^{4*}$} \\
\textbf{Samuel Nyarko$^{15*}$, Cynthia Amoaba$^{*}$, Esther Dansoa Appiah$^{8*}$, Felix Akwerh$^{2*}$} \\
\textbf{Richard Nii Lante Lawson$^{9*}$, Joel Budu$^{10*}$, Emmanuel Debrah$^{4*}$, Nana Boateng$^{1*}$} \\
\textbf{Wisdom Ofori$^{*}$, Edwin Buabeng-Munkoh$^{*}$, Franklin Adjei$^{11*}$, Isaac K. E. Ampomah$^{12*}$} \\
\textbf{Joseph Otoo$^{13*}$, Reindorf Borkor$^{2*}$, Standylove Birago Mensah$^{2*}$, Lucien Mensah$^{7*}$} \\
\textbf{Mark Amoako Marcel$^{*}$, Anokye Acheampong Amponsah$^{14*}$, and James Ben Hayfron-Acquah$^{2*}.$} \\ \\

 $^*$ NLP Ghana,$^1$ Algorine,$^2$ Kwame Nkrumah University of Science and Technology, \\
 $^3$ Leuphana University Luneburg,$^4$Department of Mathematics, University of Cape Coast,\\  $^5$ Edinburgh Napier University, $^6$ Accra Institute of Technology,$^7$ Tulane University,\\ $^{8}$ University of Tromso, $^{9}$ AiMlCamp,  $^{10}$ University of Strathclyde, $^{11}$ Azubi Africa,\\ $^{12}$ Ulster University, 
 $^{13}$ Centre for Research, Data Science and IT Solutions, \\ $^{14}$ University of Energy and Natural Resources,\\ $^{15}$ Integrated Geospatial Intelligence Application Centre, SRH Berlin University of Applied Science.
  }

\date{}

\begin{document}
\maketitle
\begin{abstract}
Transformer-based language models have been changing the modern Natural Language Processing (NLP) landscape for high-resource languages such as English, Chinese, Russian, etc. However, this technology does not yet exist for any Ghanaian language. In this paper, we introduce the first of such models for Twi or Akan, the most widely spoken Ghanaian language. The specific contribution of this research work is the development of several pretrained transformer language models for the Akuapem and Asante dialects of Twi, paving the way for advances in application areas such as Named Entity Recognition (NER), Neural Machine Translation (NMT), Sentiment Analysis (SA) and Part-of-Speech (POS) tagging. Specifically, we introduce four different flavours of ABENA -- {\it A BERT model Now in Akan} that is fine-tuned on a set of Akan corpora, and BAKO - BERT with Akan Knowledge only, which is trained from scratch. We open-source the model through the Hugging Face model hub and demonstrate its use via a simple sentiment classification example.

\end{abstract}

\section{Introduction}
Natural Language Processing (NLP) is a subfield of artificial intelligence that is primarily concerned with developing systems that can read and process spoken or written human language to perform intelligent tasks. In order to capture the meaning encoded in a sentence or a phrase, a language model can be employed to extract such information. 

A language model is a core component of NLP, accurately placing distributions over sentences to decode the complexities of a language -- such as grammatical structure and semantics ~\citep{jozefowicz2016exploring}. Over the years, there have been many attempts at developing language models that fully capture the context of a sentence and also better models for the observed distributions of a language. Recent advances in transformer ~\citep{vaswani2017attention} neural language models such as BERT ~\citep{BERT} and GPT ~\citep{GPT1} ~\citep{GPT2} has stimulated significant progress in performance for high-resource languages such as English, Chinese, etc. These models are typically trained on billions of words. While text data and other computational resources are easily accessible for some languages, text data resources in many low-resource languages remain almost non-existent. In this work, we contribute to closing this gap by developing the first transformer language model for the low-resource Ghanaian language Twi or Akan.

Human languages share some similarities in both acoustic and phonetic aspects. Features extracted from some languages can be shared with other languages at some levels of abstraction ~\citep{chung2018unsupervised, jia2020large}. Inspired by that, the main idea is to first train a high-resource language model, and then use the resulting trained network (the {\it parent} or {\it source} model) to initialize and fine-tune a low-resource language model (the {\it child} or {\it target} model) ~\citep{nguyen2017transfer}. This process is typically referred to as \textit{transfer learning} ~\citep{RuderTransferLearning,AzunreTransferLearning}.

In this paper, we develop several pretrained transformer language models for the Akuapem and Asante dialects of Twi, paving the way for advances in application areas such as Named Entity Recognition (NER), Neural Machine Translation (NMT), Sentiment Analysis (SA) and Part-of-Speech (POS) tagging. We introduce four different flavors of ABENA -- {\it A BERT model Now in Akan}. We first employ transfer learning to fine-tune a multilingual BERT (mBERT) model on the Akuapem Twi subset of the JW300 dataset. Subsequently, we fine-tune this model further on the Asante Twi Bible data to obtain an Asante Twi version of the model. Additionally, we perform both experiments using the DistilBERT architecture instead of BERT. This yields smaller and more lightweight versions of the Akuapem and Asante ABENA models. Finally, we introduce BAKO -- {\it BERT with Akan Knowledge Only}. This is essentially the same as ABENA, but trained from scratch. We open-source the model through the \href{https://huggingface.co/Ghana-NLP}{Hugging Face model hub}, and demonstrate its use via a simple sentiment classification example.

\section{Related Work}
Global matrix factorization methods, such as Latent Semantic Analysis (LSA)~\citep{deerwester1990indexing} learn vector space representation of words and transform textual data into numerical representations which uses mathematical models to learn dependencies between each word, phrase and document ~\citep{mandygu}. The process of turning text into numbers is commonly known as \textit{vectorization or word embedding}. These techniques are functions which map words into vectors of real numbers. The vectors form a vector space -- an algebraic model where all the rules of vector addition and measures of similarities apply. 

\subsection{Language Modeling}
Language modeling is formulated as the task of predicting the $(n+1)^{th}$ token in a sequence given the $n$ preceding tokens ~\citep{merity2018analysis}. Mathematically, a language model may be expressed as $P(x^{n+1} \mid x^{n},...., x^{1})$
where
$P(x^{T}, x^{T-1},...,x^{1})  =  P(x^{1})\cdot P(x^{2} \mid x^{1})\times....\times P(x^{T} \mid x^{T-1},...x^{1})
= \prod_{t=1}^{T} P(x^{t} \mid x^{t-1},...,x^{1}).$

Language models can operate at various granularities, with tokens formed from either words, sub-words, or characters ~\citep{merity2018analysis} which can be used to accurately predict the next word in a sentence. Further, when trained on vast amounts of data, language models compactly extract meaning encoded in the training data ~\citep{jozefowicz2016exploring}. 

Primarily, there are two types of language models: statistical language models and neural language models. The difference between the two lies in their mode of operation and architecture. 

\subsubsection{Statistical Language Models}
Statistical language models are based on the concepts of assigning a probability distribution to words and sentences ~\citep{popov2018catalytic} where each sentence is an ordered sequence of words. Every word $w$ is taken from a set $D$ also called a \textit{dictionary}. 

The goal of a statistical language model is to build a probabilistic distribution over all possible sentences in a language ~\citep{raychev2014code}. That is, given a sentence $S$, the language model estimates its probability $P(S)$. 

For a sentence $S = w_{1}, w_{2},...,w_{m}$, many language modeling approaches estimate its probability as follows:  
\[ P(S) = \prod P(w_{i} \mid h_{i-1}) \quad \forall i : \quad 1 \leq i \leq m,\] where we refer to the sequence $h_{i}=w_{1}.w_{2}.,..w_{i}$ as the history. A sentence likelihood is calculated by generating it word by word using conditional probabilities on the already generated words. Furthermore, a statistical language model is usually built on a finite amount of training data that is used to estimate the true probabilities of sentences. 

There are several variants of the statistical language model, like the \textit{n-gram} models and count-based models. Almost all these variants of the statistical language model decompose the probability of a sentence into a product of conditional probabilities, i.e., instead of modeling the probability of each sentence independently, the probability is composed of probabilities of its sub-parts or words. 

Statistical language models are also regarded as knowledge-improvised data-optimal techniques ~\citep{rosenfeld2000two} which fail to capture long-term dependencies and linguistic structure in their architecture. Owing to this, statistical language models inherently suffer from data sparsity. Some solutions have been exploited like the Good-Turing~\citep{church1991comparison}, Kneyser-Ney~\citep{heafield2013scalable} and the Witten-bell ~\citep{giwa2013n} algorithms. Aside these, domain specific and domain adaptive statistical language models have been used~\citep{rosenfeld1996maximum}, but these solutions are computationally expensive and still fail to fully capture context in sentences with long-term dependencies.

Statistical language models also have a fundamental problem known as the \textit{curse of dimensionality} which limits modeling on larger corpora for a universal language model ~\citep{ranjan2016survey}. For example, if one wants to develop an n-gram language model with a vocabulary of size $1000$, there are potentially $1000^{n} - 1$ possible dimensions in the resulting vector space.

In view of these challenges, we turn our attention to neural language models.

\subsubsection{Neural Language Models - Static}
Neural Networks are a class of machine learning algorithms that learns to perform a task by modeling probabilities using continuous internal representations of the input feature. Some of these network architectures are employed in language modeling, and the resulting models are known as \textit{neural language models}~\citep{ahn2016neural}. Neural language models~\citep{xu2000can} traditionally used convolutional neural networks (CNNs)~\citep{o2015introduction} and recurrent neural networks (RNNs)~\citep{mikolov2010recurrent} -- like LSTMs~\citep{hochreiter1997long,gers2002learning,sundermeyer2012lstm} and GRUs~\citep{cho2014properties,chung2014empirical} -- with transformers becoming more popular recently. 

Unlike statistical language models,the conditional probabilites -- $P(x_{t} \mid x_{1:t-1})$ -- can be specified using a neural network in which the context $x_{1:t-1}$ at each time $t$ is represented using a hidden state vector $h_t \in \mathbb{R}^d $. This is defined recursively via $h_t = f(x_{t-1},h_{t-1};\theta)$, where $f$ is a non linear map with a trainable parameter $\theta$. The conditional probabilites are then defined using a softmax function as $P(x_t \mid x_{1:t-1}:\theta,\omega)$ = softmax $(X_t,\omega,h_t)$ where $\omega ={\omega_i}\subset \mathbb{R}^d$ is the softmax coefficient. $\omega_i$ can be viewed as an embedding vector for word $i \in V$ and $h_t$ the embedding vector of context $x_{1:t-1}$ which is converted into a probability using the softmax function.

Key to neural language modeling are word embeddings which represents words in the form of vectors~\citep{chen2018word2vec}. It is a form of learning representation which is primarily concerned with representing each word using an $N$-dimensional vector such that the cosine similarity of any two terms corresponds to their semantic similarity~\citep{speer2016ensemble}.
The next section discusses some techniques that are used to learn word embedding from text data.\\

\textbf{Word2vec}\\
Word2vec is a method of unsupervised shallow two-layer neural network modeling ~\citep{goldberg2014word2vec} that captures word embeddings from a text corpus. Word2vec produces embeddings based on the contextual semantics of words in a text such that words with similar linguistic contexts are mathematically grouped together in a vector space, preserving the semantic relationship between words. It then uses these word embeddings to produce predictions on a word. 

There are two model architectures of word2vec that can be used:

\begin{itemize}
\item Skip-Gram  Model 
\item Continuous Bag-of-Words Model (CBOW)
\end{itemize}

\textbf{FastText}\\
FastText embeddings exploit subword information to construct word embeddings. Representation are learnt of character $n$-grams and words represented as the sum of the $n$-grams vectors~\citep{bojanowski2017enriching}. This extends the word2vec type of models with subword information and helps the embeddings understand suffixes and prefixes. Once a word is represented using character $n$-grams, a skip-gram model is trained to learn the embeddings.\\

\textbf{Sent2vec}\\
Sent2vec is an efficient unsupervised word embedding architecture that seeks to train distributed representations of sentences~\citep{pagliardini2017unsupervised}. The model is further augmented by also learning source embeddings for not only unigrams but also $n$-grams of words present in each sentence, and averaging them along with the words. It can be thought of as an extension of fastText and word2vec.\\

\textbf{Doc2vec}\\
Doc2vec ~\citep{lau2016empirical} is an extension of word2vec. The model aims to create a numerical representation of a document rather than individual words~\citep{chen2018word2vec}. It functions on the logic that the meaning of a word is largely dependent on the document that it occurs in. The vectors generated by doc2vec are generally used to find the similarities between documents.

\subsubsection{Neural Language Models - Contextual}
Unlike the static embeddings like word2vec and fastText, transformer-based language models have the ability to dynamically adapt to changing contexts. For instance, consider the following two English sentences:

\begin{itemize}
\item He examined the \textit{cell} under the microscope.
\item He was locked in a \textit{cell}.
\end{itemize}

The same word \textit{cell} here means two completely different things. A human would look at the context, i.e., the words surrounding our target word, for cues about the exact meaning. The word \textit{microscope} clearly brings up biological connotations in the first sentence. The word \textit{locked} clearly brings up connotations of a prison in the second sentence.

To make a Twi example, consider the following two sentences.

\begin{itemize}
\item \textopeno kraman no \textit{so} paa — the dog is very \textit{big}.
\item \textopeno kra no da mpa no \textit{so} — the cat is sleeping \textit{on} the bed.
\end{itemize}

The word \textit{so} means \textit{big} in the first sentence and \textit{on} in the second sentence.
Transformer-based models possess the ability to dynamically analyze the context for meaning cues, and adapt to it. They do this using the so called \textit{self-attention mechanism}, which we do not discuss in detail here. BERT is arguably the most popular model from this class. It is typically trained with a \textit{``fill-in-the-blanks”} objective which is very practical to implement and does not require labeled data — just randomly drop some words and try to predict them. It is also trained with the task of predicting if the next sentence is a plausible one to follow the current one, i.e., \textit{next sentence prediction}.

These models are equipped with the ability to be adapted or fine-tuned to new tasks, scenarios and languages. As already mentioned earlier, this adaptation process is also referred to as transfer learning. Nowadays, it is a lot less common to train any model from scratch — instead, models trained to fill-in-the blanks (and other tasks) on billions of words are shared openly and fine-tuned to new scenarios using very little additional data. This is more representative of how humans learn, as we typically make associations to what we already know when faced with new challenges.

Since BERT is a very large model, its large variant being approximately 179 million parameters in size, a number of recent efforts have been aimed at making them smaller and more lightweight. These include ALBERT~\citep{ALBERT}, RoBERTa~\citep{RoBERTa} and DistilBERT ~\citep{DistilBERT}. Here, we utilize DistilBERT, which employs a technique known as \textit{knowledge distillation}, to produce smaller distilled versions of ABENA. We also experiment with RoBERTa for training models from scratch.

\subsection{Static Word Embeddings for Twi}
The work by \citet{alabi2020massive} constructed static word embeddings for the Twi language. In their work, the researches made use of two main architectures which analyzed representation of words of the Twi language at both word and character levels, in order to explore the semantics of the language. The two main architectures used were fastText \citep{bojanowski2017enriching} and Character Word Embedding (CWE) \citep{chen2015joint}. They noted that there is a large amount of ambiguity in the Twi language, hence the decision to choose a CWE architecture.

The authors performed experiments which checked how these architectures performed on three different classes of corpora for the Twi language: i) a clean corpora (Asante Bible), ii) clean plus some noisy corpora (Asante Bible and Twi Wikipedia) and iii) all clean and all noisy corpora (Bible, Wikipedia and JW300 \citep{agic2020jw300}). For evaluation, researchers and linguists translated the \textit{wordsim-353} \citep{finkelstein2001placing} dataset from English to Twi, with the resulting dataset being a notable contribution. The Spearman $\rho$ correlation between human judgement encoded in this dataset and cosine similarity scores on the wordsim-353 dataset were used as the metric to evaluate these embeddings.

It was discovered that the ``clean plus some noisy" Twi data gave the best result, as evident by the best correlation with human judgements $(\rho^{C2}_{CWE} = 0.437$  vs.  $\rho^{C2}_{fastText} = 0.388)$. As a precursor to the work we report in this paper, we first worked on replicating the fastText component of their results\footnote{\href {https://nlpghana.medium.com/introducing-abena-bert-natural-language-processing-for-twi-d55a6cb312ee}{Introducing ABENA (Medium article)}}. While results from training directly on their data was comparable, we were able to improve on the value of $\rho^{C2}_{fastText}$ by training on the large noisy JW300 corpus only. This yielded a $\rho$ value of $0.421$ that is close to the value they achieved with CWE.

\section{Dataset}
JW300 is a parallel corpus of over 300 languages with around 100 thousand parallel sentences per language pair on average. It spans across 343 languages, and comprises a total of 1,335,376 articles -- with a bit over 109 million sentences, and 1.48 billion tokens \citep{agic2020jw300}. It is a thorough mine of all the publications from the \textit{jw.org} website. A vast majority of texts come from the magazines \textit{Watchtower} and \textit{Awake!} Although the articles emanate from a religious society, they encompass a various range of topics. 

JW300 plays a vital role in low-resource language coverage \citep{orife2020masakhane}. For example, OPUS \citep{tiedemann-2012-parallel}, another massive parallel text collection, offers over 100 million English-German parallel sentences, and JW300 only 2.1 million. However, in another example (for Afrikaans to Croatian) the counts are 300 thousand in OPUS and 990 thousand in JW300. The basic statistics of JW300 reveal up to 2.5 million sentences for high-resource languages such as English, French, and Italian. However, the long tail of low-resource languages typically still offers between 50-100 thousand sentences. In comparison to the OPUS corpus, JW300 fills an important gap in cross-lingual resources: it comprises a multitude of low-resource languages while still offering ample sentences for each individual language, as well as parallel sentences for language pairs. 

Contrary to the positive aforementioned details, the models trained with this dataset naturally show a varying degree of religious bias. The JW300 dataset is freely available for all non-commercial use.

\section{ABENA Training Experiment}
First, we initialize a BERT neural architecture and corresponding tokenizer to the multilingual BERT (mBERT) checkpoint\footnote{ \href{https://github.com/google-research/bert}{BERT github repo}}. This model checkpoint was trained on over $100$ languages simultaneously. Although these did not include any Ghanaian languages, it does include another ``Niger-Congo” language — Nigerian Yoruba. It is thus reasonable to expect that this model contains some knowledge useful for constructing a Twi embedding. 

We transfer this knowledge by fine-tuning the initialized mBERT weights and tokenizer on the monolingual Twi subset of the JW300 corpus. The convergence info is shown in Table \ref{table:abenaCONV}. All models were trained on a single Tesla K80 GPU on a single NC6 Azure Virtual Machine instance.

\begin{table}
\centering
\begin{tabular}{lrl}
\hline \textbf{Model} & \textbf{Epochs/Time} & \textbf{Loss} \\ \hline
Akuapem (cased) & 3 epochs, 35 hrs & 0.56 \\
Asante (uncased) & 3 epochs, 1.8 hrs & 0.95\\
\hline
\end{tabular}
\caption{Convergence info for ABENA models. All models were trained on a single Tesla K80 GPU on an NC6 Azure VM instance. }
\label{table:abenaCONV}
\end{table}

The time to run $3$ epochs is significantly shorter for the Asante model because the Asante Twi Bible is significantly smaller than JW300 (25k+ samples versus 600k+ samples). Moreover, note that we further made the Asante model uncased due to this smaller training data size. This just means that the model does not distinguish between upper and lower cases of letters, while the Akuapem model does. We present the final loss value attained for all models, in the absence of any other benchmarks available for the Twi languages. In the final section we will outline some of the things we are doing to address this lack of benchmarks. Recall that the BERT loss is the sum of the ``fill-in-the-blanks" and next sentence prediction losses.

Next, we perform the same experiments using the DistilBERT architecture ~\citep{DistilBERT} instead of BERT. This yields smaller and more lightweight versions of both the Akuapem and Asante ABENA models. DistilBERT employs a technique known as \textit{knowledge distillation} to reduce the size of the model while sacrificing little of the performance. If DistilBERT is the distilled version of BERT, then DistilABENA is the distilled version of ABENA. 

The distilled models have approximately 135 million parameters, which is $75\%$ of the $179$ million parameters in the BERT-based models. Moreover, we experiment with training our own tokenizer on the monolingual Twi data from scratch prior to fine-tuning (versus using the included pretrained multilingual tokenizer as before). This yields ``V2" versions of the distilled models, with significantly higher final loss values as shown in Table \ref{table:distilabenaCONV}. We share these models as well, to make it easy for the community to further experiment with and fine-tune them. Intuitively, we expect a monolingual tokenizer to eventually be more precise, even if the quality and amount of the training data at the moment precludes us from fully demonstrating this.

\begin{table}
\centering
\begin{tabular}{lrl}
\hline \textbf{Model} & \textbf{Epochs/Time} & \textbf{Loss} \\ \hline
Akuapem & 3 epochs, 26.9 hrs & 0.65 \\
Asante ``V2" & 5 epochs, 26.7 hrs & 2.2\\
Akuapem & 3 epochs, 1.4 hrs & 1.05 \\
Asante ``V2'' & 5 epochs, 1.5 hrs & 3.5\\
\hline
\end{tabular}
\caption{Convergence info for DistilABENA models. All models were trained on a single Tesla K80 GPU on an NC6 Azure VM instance. ``V2'' models have tokenizers trained from scratch.}
\label{table:distilabenaCONV}
\end{table}

\section{BAKO Training Experiment}
We also investigated training the various forms of ABENA described in the previous section from scratch on the monolingual data. We named these set of models BAKO-\textit{``BERT with Akan Knowledge Only"}. We found BERT and DistilBERT not suitable for this right now, given the relatively small size of the JW300 dataset. For this reason we only present the RoBERTa ~\citep{RoBERTa} versions of BAKO, i.e., RoBAKO. RoBERTa is an improvement on BERT that employs some optimization tricks for better efficiency. One notable difference is that it uses \textit{byte-pair encoding} (BPE) for tokenization, versus the \textit{WordPiece} approach used by BERT and DistilBERT. Importantly, it discards the next sentence prediction task of BERT. In fact, the RoBERTa-based models have ``only" $84$ million parameters, which is just $47\%$ the size of the BERT-based models. We show the training information for these models in the Table \ref{table:robakoCONV}. 

\begin{table}
\centering
\begin{tabular}{lrl}
\hline \textbf{Model} & \textbf{Epochs/Time} & \textbf{Loss} \\ \hline
Akuapem & 5 epochs, 15.3 hrs & 2.2 \\
Asante & 5 epochs, 1 hrs & 3.5\\
\hline
\end{tabular}
\caption{Convergence info for RoBAKO models trained from scratch. All models were trained on a single Tesla K80 GPU on an NC6 Azure VM instance.}
\label{table:robakoCONV}
\end{table}

\section{Simple Examples}
\subsection{Fill-in-the-blanks}
We present two (2) example masked sentences and resulting completions by our model. Several other examples could be experimented within minutes using our Kaggle ABENA usage demo\footnote{\href {https://www.kaggle.com/azunre/ghananlp-abena-usage-demo}{ABENA usage demo on Kaggle}}.

The Python code needed to do this is shown next.

\begin{lstlisting}[language=Python]
MODEL = "Ghana-NLP/distilabena-base-akuapem-twi-cased" # Akuapem DistilABENA

from transformers import pipeline

fill_mask = pipeline(
    "fill-mask",
    model=MODEL,
    tokenizer=MODEL
)

# First Sentence
print(fill_mask("Saa tebea yi maa me papa [MASK].")) 
In English the above sentence reads ("This situation made my father [mask]")

# Second Sentence
print(fill_mask("Eyi de ohaw kese baa [MASK] ho."))
In English the above sentence reads ("This presented a difficult problem for [mask]")
\end{lstlisting}
As you can see, having hosted this on the Hugging Face model repo, the code required to do this is only a few lines. Table \ref{s1} and Table \ref{s2} display the top 5 completions for the two (2) example masked sentences.

\begin{table}
\centering
\begin{tabular}{lr}
\hline \textbf{Completed Sentence} & \textbf{Score} \\ \hline
Saa tebea yi maa me papa no &  0.228 \\
Saa tebea yi maa me papa boom  & 0.068\\
Saa tebea yi maa me papa bio & 0.057\\
Saa tebea yi maa me papaapa & 0.038\\
Saa tebea yi maa me papa da & 0.027\\
\hline
\end{tabular}
\caption{Top five completions for first example sentence}
\label{s1}
\end{table}

\begin{table}
\centering
\resizebox{\columnwidth}{!}{%
\begin{tabular}{lr}
\hline \textbf{Completed Sentence} & \textbf{Score} \\ \hline
Eyi de \textopeno haw k\textepsilon se baa me h\textopeno & 0.171 \\
(This presented a difficult problem for me) & 0.171\\
Eyi de \textopeno haw k\textepsilon se baa Adam h\textopeno & 0.149\\
(This presented a difficult problem for Adam) & 0.149\\
Eyi de \textopeno haw k\textepsilon se baa ne h\textopeno & 0.113\\
(This presented a difficult problem for him) & 0.113\\
Eyi de \textopeno haw k\textepsilon se baa fie h\textopeno & 0.053\\
(This presented a difficult problem for the house) & 0.053\\
Eyi de \textopeno haw k\textepsilon se baa Satan h\textopeno & 0.047\\
(This presents a difficult problem for Satan) & 0.047\\
\hline
\end{tabular}%
}
\caption{Top Five Completions for the Second Sentence}
\label{s2}
\end{table}
The religious bias is clearly evident. For instance, when completing the sentence “Eyi de \textopeno haw k\textepsilon se baa \rule{1cm}{0.15mm} h\textopeno”, you get completions such as “Eyi de \textopeno haw k\textepsilon se baa \textbf{Adam} h\textopeno” and/or “Eyi de \textopeno haw k\textepsilon se baa \textbf{Satan} h\textopeno” among the most likely completions. These indicate a strong level of religious bias as the training dataset emanates from a religious society. Otherwise the completions seem to be reasonable.

\subsection{Sentiment Classification}
In the context of sentiment classification, we built a simple sentiment analysis dataset of just 20 samples and experimented with the features of our model above to solve a classification problem. Detailed Python code and results of the sentiment analysis can be obtained from our ABENA usage demo\footnote{\href {https://www.kaggle.com/azunre/ghananlp-abena-usage-demo}{ABENA usage demo on Kaggle}}.

Briefly, we created a balanced dataset of 20 samples -- 10 indicating positive sentiment, such as \textit{M’ani  agye papa} (I am very happy), and 10 indicating negative sentiment, such as \textit{Sister Akosua aduane no nny\textepsilon}  \textit{d\textepsilon} (Sister Akosua's food isn't good). We train a simple kNN classifier on 14 random subsamples with \textit{sklearn} and test on the remaining 6. In our testing of many repeated runs, we found the accuracy to consistently fall in the 83\% to 100\% range, and always outperform 50\% / random chance. Although testing on a bigger dataset is required -- an effort currently underway -- these results are very promising.

\section{Discussion and Conclusion}
As we already mentioned, the models show a varying degree of religious bias. For instance when completing a sentence like ``Eyi de \textopeno haw k\textepsilon se baa \rule{1cm}{0.15mm} h\textopeno", you may see completions such as ``Eyi de \textopeno haw k\textepsilon se baa \textbf{Adam} h\textopeno" and/or ``Eyi de \textopeno haw k\textepsilon se baa \textbf{Satan} h\textopeno" among the most likely completions. While these are not technically wrong -- they can be useful for understanding sentence structure, part of speech likely to appear in that location, etc. -- the fact that these are among the top $5$ completions indicates a strong religious bias in the model. This is obviously at least partly due to the religious bias in the data used to train/fine-tune them.

This is the best we could do at the moment, due to the extreme low-resource nature of Twi, and it is clearly already a useful set of models. However, it is very important to communicate that this is a work in progress, and the models should be used with care and awareness of their religious bias. 

In an ongoing attempt to reduce the religious bias in the JW300 dataset to a minimum, we have recently curated an English to Akuapem Twi parallel machine translation corpus with 25,421 sentence pairs\footnote{\href {https://zenodo.org/record/4432117}{Akuapem to English Dataset}}. The purpose of this dataset is to fine-tune the model further after training it on the JW300 dataset. We are in the process of creating even larger datasets on which we can fine-tune these models further to reduce the bias.

The work we have presented opens up a number of research directions for us to pursue in order to further improve these models, and enable many critical applications. In no particular order, the following list describes these research efforts, some of which are already ongoing.
\begin{enumerate}
    \item Fine-tuning on cleaner, higher-quality and less-biased Twi data
    \item Developing a Twi version of the labeled GLUE Dataset ~\citet{GLUE} for fine-tuning on many other tasks, such as entailment, sentiment analysis, sentence similarity, etc.
    \item Training ALBERT ~\citet{ALBERT} for Twi and comparing with presented models. This architecture promises an even greater size saving than RoBERTa.
    \item Unsupervised spell-checking methods based on these models
    \item Unsupervised Named Entity Recognition (NER) methods based on these models
    \item Developing a Twi version of the GPT ~\citet{GPT1}~\citet{GPT2} causal transfomer-based text generator
    \item Developing a labeled Named Entity Recognition (NER) dataset, e.g., a translation of CONLL 2003, and fine-tuning these models on it
    \item Incorporating all of the above developments into an open Python API to make all of these Ghanaian NLP tools readily available to NLP developers and practitioners
    \item Extending all of the above to other overlooked Ghanaian and African languages
\end{enumerate}

\section*{Acknowledgments}

We are grateful to the Microsoft for Startups Social Impact Program -- for supporting this research effort via providing GPU compute through Algorine Research. We would like to thank Julia Kreutzer, Jade Abbot, Emmanuel Agbeli and Deborah Kanubala for their constructive feedback. We are grateful to the reviewers for their valuable comments. We would also like to thank the \href{https://gajreport.com}{Ghanaian American Journal} for their work in sharing our work and mission with the Ghanaian public.

\bibliography{anthology,eacl2021}
\bibliographystyle{acl_natbib}

\end{document}